\def\year{2021}\relax
\documentclass[letterpaper]{article} % DO NOT CHANGE THIS
\usepackage{aaai21}  % DO NOT CHANGE THIS
\usepackage{times}  % DO NOT CHANGE THIS
\usepackage{helvet} % DO NOT CHANGE THIS
\usepackage{courier}  % DO NOT CHANGE THIS
\usepackage[hyphens]{url}  % DO NOT CHANGE THIS
\usepackage{graphicx} % DO NOT CHANGE THIS
\usepackage{natbib}  % DO NOT CHANGE THIS AND DO NOT ADD ANY OPTIONS TO IT
\usepackage{caption} % DO NOT CHANGE THIS AND DO NOT ADD ANY OPTIONS TO IT
\usepackage{wrapfig}
\usepackage{amsmath}
\usepackage{amsthm}
\usepackage{amssymb}
\usepackage[switch]{lineno}

\theoremstyle{definition}
\newtheorem{definition}{Definition}

\usepackage{subcaption}
\usepackage{caption}

\title{Deep Stock Trading: A Hierarchical Reinforcement Learning Framework for Portfolio Optimization and Order Execution\footnote{This paper was accepted by AAAI 2021 with the original title: Commission Fee is not Enough: A Hierarchical Reinforced Framework for Portfolio Management}}

\author{

Rundong Wang\textsuperscript{\rm 1}\thanks{Equal contribution},
Hongxin Wei\textsuperscript{\rm 1}\footnotemark[1],
Bo An\textsuperscript{\rm 1},
Zhouyan Feng\textsuperscript{\rm 2}, 
Jun Yao\textsuperscript{\rm 2}\\
    
}
\affiliations{
    \textsuperscript{\rm 1}School of Computer Science and Engineering, Nanyang Technological University, Singapore\\
    \textsuperscript{\rm 2} WeBank Co. Ltd., China\\
\{rundong001, hongxin001\}@e.ntu.edu.sg, boan@ntu.edu.sg, \{yanisfeng, junyao\}@webank.com

}
\begin{document}
% \linenumbers

\maketitle

\begin{abstract}
Portfolio management via reinforcement learning is at the forefront of fintech research, which explores how to optimally reallocate a fund into different financial assets over the long term by trial-and-error. Existing methods are impractical since they usually assume each reallocation can be finished immediately and thus ignoring the price slippage as part of the trading cost. To address these issues, we propose a hierarchical reinforced stock trading system for portfolio management (\textit{HRPM}). Concretely, we decompose the trading process into a hierarchy of portfolio management over trade execution and train the corresponding policies. The high-level policy gives portfolio weights at a lower frequency to maximize the long term profit and invokes the low-level policy to sell or buy the corresponding shares within a short time window at a higher frequency to minimize the trading cost. We train two levels of policies via pre-training scheme and iterative training scheme for data efficiency. Extensive experimental results in the U.S. market and the China market demonstrate that \textit{HRPM} achieves significant improvement against many state-of-the-art approaches. 
\end{abstract}

\vspace{-20pt}

\section{Introduction}

% which remains two unsolved issues: (1) \textit{non-zero slippage} - portfolio weights cannot change immediately since the trades take times to be executed; (2) \textit{transaction cost} - trading strategies which use large numbers of bids and asks can result in very large cost in the real-world

The problem of portfolio management is widely studied in the area of algorithmic trading. It aims to maximize the expected returns of multiple risky assets. Recently, reinforcement learning (RL) models are gaining popularity in the fintech community \cite{mosavi2020comprehensive}, with its great performance in different fields, including playing games \cite{silver2016mastering, mnih2013playing}, controlling robots \cite{lillicrap2015continuous}, and the Internet of things \cite{zhang2020empowering}.

Due to the dynamic nature of markets and noisy financial data, RL has been proposed as a suitable candidate by its own nature \cite{almahdi2017adaptive,jiang2017cryptocurrency}. That is, RL can directly output trade positions, bypassing the explicit forecasting step by the trial-and-error interaction with the financial environment. 
Many existing RL methods get promising results by focusing on various technologies to extract richer representation, e.g., by model-based learning \cite{tang2018actor,yu2019model}, by adversarial learning \cite{liang2018adversarial}, or by state augmentation \cite{ye2020reinforcement}. However, these RL algorithms assume that portfolio weights can change immediately at the last price once an order is placed. This assumption leads to a non-negligible trading cost, that is, \textit{the price slippage} -- the relative deviation of the expected price of a trade and the price actually achieved. In a more realistic trading environment, the trading cost should be taken into consideration. Moreover, due to the need of balancing the long-term profit maximization and short-term trade execution, it is challenge for a single/flat RL algorithm to operate on different levels of temporal tasks.

As our motivation, we observe that there is a hierarchy of portfolio managers and traders in real-world trading. Portfolio managers assign a percentage weighting to every stock in the portfolio periodically for a long-term profit, while traders care about the best execution at the favorable price to minimize the trading cost. This paper gives the first attempt to leverage a similar idea to algorithmic trading. Concretely, we develop a \textbf{H}ierarchical \textbf{R}einforced trading system for \textbf{P}ortfolio \textbf{M}anagement (\textit{HRPM}). Our system consists of a hierarchy of two decision processes. The high-level policy changes portfolio weights at a lower frequency, while the low-level policy decides at which price and what quantity to place the bid or ask orders within a short time window to fulfill goals from the high-level policy. 

Our contributions are four-fold. First, we formulate the portfolio management problem with trading cost as a hierarchical MDP, and combine portfolio management and trade execution by utilizing the hierarchical RL (HRL) framework to address the non-zero slippage. Second, as traditional HRL performs poorly in this problem, we propose the entropy bonus in the high-level policy to avoid risk, and use action branching to handle multi-dimensional action space. Third, to make full use of the stock data, we pre-train multiple low-level policies with an iterative scheme for different stocks in the limit order book environment, and then train a high-level policy on the top of the low-level policies. Fourth, we compare \textit{HRPM} with baselines based on the historical price data and limit order data from the U.S. market and the China market. Extensive experimental results demonstrate that \textit{HRPM} achieves significant improvement against state-of-the-art approaches. Furthermore, the ablation studies demonstrate the effectiveness of entropy bonuses and the trading cost is a non-negligible part in portfolio management.

\section{Problem Formulation}

\begin{figure}[!htb]
  \centering
    {
       \includegraphics[width=0.47\textwidth]{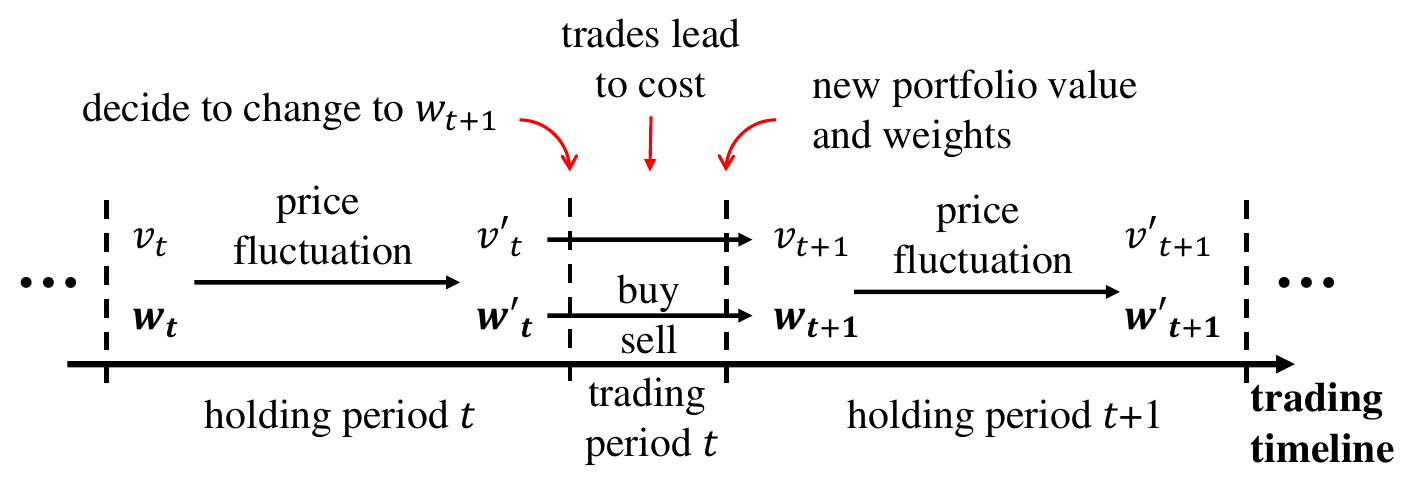}
    }
    \caption{Illustration of the portfolio management process.}
    \label{fig:the portfolio management process}
\end{figure}

In this section, we introduce some definitions and the objective. Portfolio management is a fundamental financial task, where investors hold a number of financial assets, e.g., stocks, bonds, as well as cash, and reallocate them periodically to maximize future profit. We split the time into two types of periods: holding period and trading period as Fig. \ref{fig:the portfolio management process}. During the holding period, the agent holds the pre-selected assets without making any purchase or selling. With the fluctuations of the market, the assets' prices would change during the holding period. At the end of the holding period, the agent will decide the new portfolio weights of the next holding period. In the trading period, the agent buys or sells some shares of assets to achieve the new portfolio weights. The lengths of the holding period and trading period are based on specific settings and can change over time. 

% Note that, even though the lengths of both periods are not fixed in this general formulation, we fix both for simplicity in our method.

\theoremstyle{definition}
\begin{definition}{(Portfolio)}
A portfolio can be represented as:
\begin{small}
\begin{equation}
    \boldsymbol{w}_{t}=\left[w_{0,t}, w_{1, t}, w_{2, t},  \dots, w_{M, t}\right]^{T} \in \mathbb{R}^{M+1} \text { and } \sum_{i=0}^{M} w_{i, t}=1
\label{eq:portfolio weights}
\end{equation}
\end{small}\normalsize
where $M+1$ is the number of portfolio's constituents, including one risk-free asset, i.e., cash, and $M$ risky assets, i.e., stock\footnote{In this paper, we focus on the stocks for ease of explanation. This framework is applicable to other kinds of assets.}. $w_{i,t}$ represents the ratio of the total portfolio value (money) invested at the beginning of the holding period $t$ on asset $i$. Specifically, $w_{0,t}$ represents the cash in hand. Also, we define $\boldsymbol{w'_{t}}$ as the portfolio weights at the end of the holding period $t$.
\end{definition}

\theoremstyle{definition}
\begin{definition}{(Asset Price)}
The opening prices $\boldsymbol{p}_{t}$ and the closing prices $\boldsymbol{p}'_{t}$ of all assets for the holding period $t$ are defined respectively as $\boldsymbol{p}_{t} = \left[p_{0,t}, p_{1, t}, p_{2, t},  \dots, p_{M, t}\right]^{T}$ and $\boldsymbol{p}'_{t} = \left[p'_{0,t}, p'_{1, t}, p'_{2, t},  \dots, p'_{M, t}\right]^{T}$. Note that the price of cash is constant, i.e., $p_{0,t}=p'_{0,t}=p_{0,t+1} \text{ for } t \ge 0$.
\end{definition}

\theoremstyle{definition}
\begin{definition}{(Portfolio Value)}
We define $v_{t}$ and $v'_{t}$ as portfolio value at the beginning and end of the holding period $t$. So we can get the change of portfolio value during the holding period $t$ and the change of portfolio weights:
\begin{small}
\begin{equation}
  v'_{t}=v_{t} \sum_{i=0}^M\frac{w_{i,t} p^{\prime}_{i,t}}{p_{i,t}} \text{ and }   {w}_{i, t}^{\prime}=\frac{\frac{w_{i,t} p^{\prime}_{i,t}}{p_{i,t}}}{\sum_{i=0}^M\frac{w_{i,t} p^{\prime}_{i,t}}{p_{i,t}}} \text{ for } i \in [0,M]
  \label{eq:value transistion}
\end{equation}
\end{small}\normalsize
\end{definition}

% Similarly, the change of portfolio weights during holding period $t$ is:
% \begin{equation}
%   {w}_{i, t}^{\prime}=\frac{\frac{w_{i,t} p^{\prime}_{i,t}}{p_{i,t}}}{\sum_{i=0}^M\frac{w_{i,t} p^{\prime}_{i,t}}{p_{i,t}}} \text{ for } i \in [0,M]
%   \label{eq:weight transistion}
% \end{equation}

% \theoremstyle{definition}
% \begin{definition}{(Price Relative Vector)}
% We define the \textit{price relative vector} of the trading period $t$ as $\boldsymbol{y}_{t}:=\left[1, \frac{p^{\prime}_{1, t}}{p^{\prime}_{1, t-1}}, \frac{p^{\prime}_{2, t}}{p^{\prime}_{2, t-1}}, \dots, \frac{p^{\prime}_{M, t}}{p^{\prime}_{M, t-1}}\right]^{T} \in \mathbb{R}^{M+1}$. Since $p^{\prime}_{0}$ represents the cash, the first element of $\boldsymbol{y}_{t}$ equals to 1. 
% \end{definition}

% % The elements of $\boldsymbol{y}_{t}$ represent the relative fluctuation of prices in the holding period $t$. The price relative vector is used to calculate the change in portfolio value during a time step -- a holding period and a trading period. 
% We assume that the market is continuous, i.e., the closing prices $p'_t$ for holding period $t$ equal to the opening prices $p_{t+1}$ for holding period $t+1$. The assumption holds if the holding period is set to start or end at a few minutes after the opening. So we can use the price relative vector to compute the change of portfolio value.

\theoremstyle{definition}
\begin{definition}{(Market order)}
\label{lo}
A market order refers to an attempt to buy or sell a stock at the current market price, expressing the desire to buy (or sell) at the best available price.
\end{definition}

\theoremstyle{definition}
\begin{definition}{(Limit Order)}
\label{lo}
A limit order is an order placed to buy or sell a number of shares at a specified price during a specified time frame. It can be modeled as a tuple $(p_{target}, \pm q_{target})$, where $q_{target}$ represents the submitted target price, $q_{target}$ represents the submitted target quantity, and $\pm$ represents trading direction (buy/sell).
\end{definition}

\theoremstyle{definition}
\begin{definition}{(Limit Order Book)}
\label{lob}
A limit order book (LOB) is a list containing all the information about the current limit orders. 
\end{definition}

We observe that a strategy pursuing a better selling price is always associated with a longer (expected) time-to-execution. However, the long time-to-execution could be very costly, since a strategy might have to trade at a bad price at the end of the trading period, especially when the market price moves downward.

\theoremstyle{definition}
\begin{definition}{(Trading Cost)}
\label{cost}
Trading cost consists of commission fee and slippage: $c_{trade} = c_{com}+c_{slippage}$. Commissions are fees that brokers charge to implement trades and computed as $c_{com} = \lambda \sum_{i=1}^M{(q_{i, target} \times p_{i,avg})}$, where $\lambda$ is the constant commission rate for both selling and buying. Slippage is the difference between the expected price of a trade and the price actually achieved. Here we define the slippage as the average execution price $p_{avg}$ achieved by the strategy relative to the price at the end of the trading period $t$: $c_{slippage} = (p_{avg} - p_{t+1}) \times (\pm q_{target})$.
\end{definition}

% \theoremstyle{definition}
% \begin{definition}{(Actual Portfolio Return)}
% Actual portfolio return is the difference between the actual ending portfolio value and the value that was required to acquire the portfolio minus all fees corresponding to the transaction. Mathematically, this is:
% \begin{equation}
%   v'_{t}=v_{t} \sum_{i=0}^M \left(\frac{w_{i,t}}{p_{i,t}} \times p^{\prime}_{i,t}\right)
%   \label{eq:value transistion}
% \end{equation}
% \end{definition}

Considering the trading cost, the change of portfolio value during a holding period and a trading period $t$ satisfies:
\begin{equation}
\begin{aligned}
v_{t+1} &= \underbrace{v_t w_{0,t} - \sum_{i=1}^M{[\pm q_{i, target} \times p_{i,avg}}+\lambda q_{i, target} \times p_{i,avg}]}_{\text{Cash}}\\ &+ \underbrace{\sum_{i=1}^M{\left[(\frac{v_t w_{i,t}}{p_{i,t}} \pm q_{i,target}) \times p_{i,t+1}\right]}}_{\text{Assets Value}} \\&= \sum_{i=0}^M{\frac{v_t w_{i,t} p_{i,t+1}}{p_{i,t}}} -c_{trade,t}
\label{eq:with transaction cost}
\end{aligned}
\end{equation}

Our objective is to maximize the final portfolio value given a long time horizon by taking into account the trading cost.

\begin{figure*}[t]
     \centering
    \begin{subfigure}{0.63\textwidth}
      \centering
      % include first image
      \includegraphics[width=\textwidth]{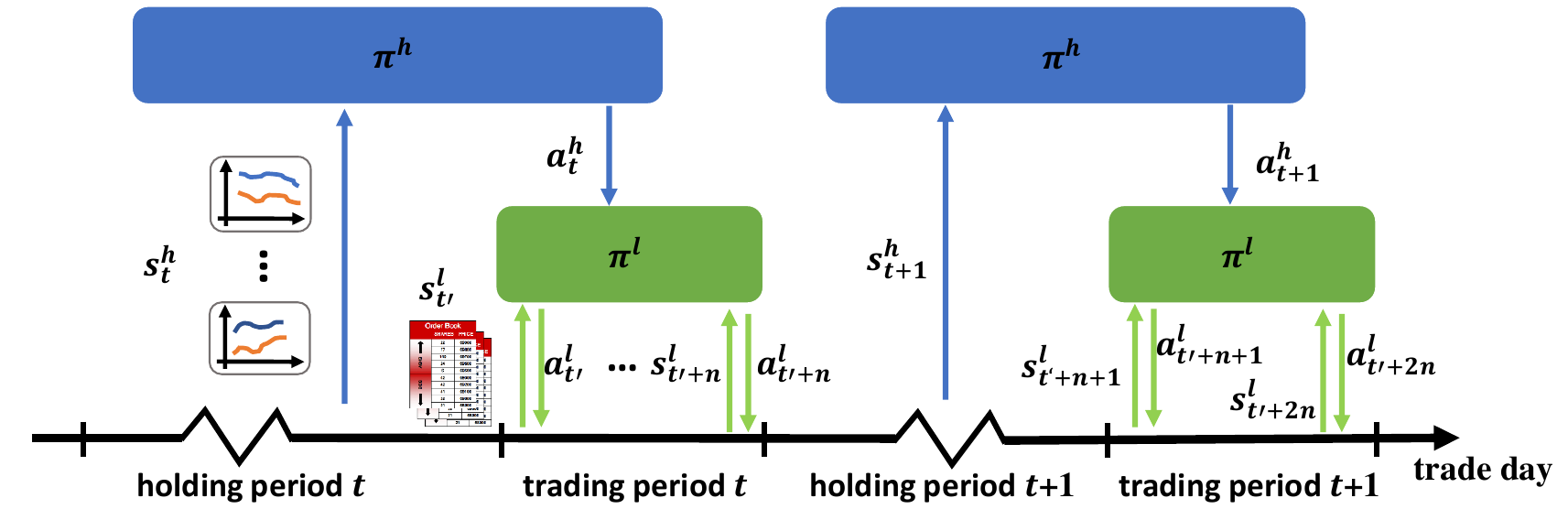}  
      \caption{\textit{HRPM} Model}
      \label{fig:hrpm}
    \end{subfigure}
    \begin{subfigure}{0.18\textwidth}
      \centering
      % include second image
      \includegraphics[width=\textwidth]{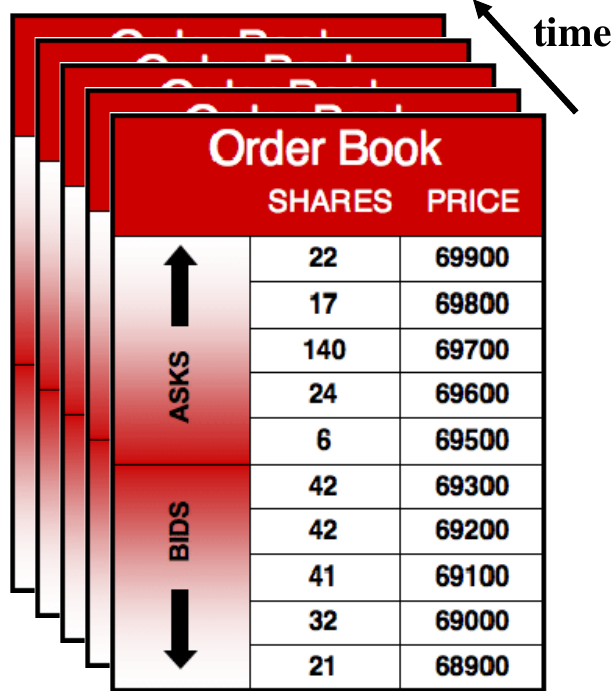}  
      \caption{The low-level states}
      \label{fig:LOBs}
    \end{subfigure}
\caption{Overview of \textit{HRPM}}
\label{fig:overview}
\end{figure*}

\section{Hierarchical MDP Framework}

In this section, we propose a hierarchical MDP framework to handle the real-world trading scenarios with slippage. We start by introducing the Markov decision process (MDP), which is defined by the tuple: $\mathrm{MDP}=\left(\mathcal{S}, \mathcal{A}, \mathcal{P}, r, \gamma, T\right)$, where $\mathcal{S}$ is a state space, $\mathcal{A}$ is an action space, $\mathcal{P}: \mathcal{S} \times \mathcal{A} \times \mathcal{S} \rightarrow \mathbb{R}_{+}$ is a transition probability function, $r: \mathcal{S} \rightarrow \mathbb{R}$ is a bounded reward function, $\gamma \in(0,1]$ is a discount factor and $T$ is a time horizon. In an MDP, an agent at state $s_{t} \in \mathcal{S}$ performs an action $a_{t} \in \mathcal{A}$. The agent's actions are often defined by a policy $\pi_{\theta}: S \rightarrow \mathcal{A}$ parameterized by $\theta$. A Q-value function gives expected accumulated reward when executing action $a_t$ in state $s_t$ and following policy $\pi$ in the future, which is $Q^\pi(s_t, a_t) = \mathbb{E}_{(s_{t+1},\cdots \sim \pi)}\left[\sum_{i=t}^{T} \gamma^{i} r(s_i,a_i)\right]$. The objective of the agent is to learn an optimal policy: $\pi_{\theta^{*}}=\operatorname{argmax}_{\pi_{\theta}} \mathbb{E}_{\pi_{\theta}}\left[\sum_{i=0}^{T} \gamma^{i} r_{t+i} | s_{t}=s\right]$.

A key challenge to the real-world portfolio management is to balance the multifaceted and sometimes conflicting objectives of different decision processes. Portfolio managers' main concerns are about longer-term profit. They typically think of execution as a chore to fulfill the clients' profit requirements and risk preferences. While traders care about best execution at the most favorable price to minimize the trading cost, but might not always be aware of managers' sources of profit and urgency of trade. Moreover, flat RL cannot handle complex tasks with long time horizons. Portfolio managers take many hours (sometimes days) while agents need to make decisions every few seconds or faster. The sampling frequency is limited by this time horizon issue so that all available information about market dynamics is not fully utilized. Consequently, electronic trading algorithms are supposed to operate on multiple levels of granularity.

Formally, we model the portfolio management with trading cost as two hierarchical MDPs, respectively for portfolio management and trade execution:
$$\operatorname{MDP}^{h} =(\mathcal{S}^h, \mathcal{A}^{h}, \mathcal{P}^h, r^{h}, \gamma, T^h)$$
$$\operatorname{MDP}^{l} =(\mathcal{S}^l, \mathcal{A}^{l}, \mathcal{P}^l, r^{l}, \gamma, T^l)$$
where the detailed definitions are given in the following subsections. The high-level policy and the low-level policy are executed in different timescales. Concretely, the time horizon for the high-level MDP is the total holding time of the portfolio, and the timestep of the high-level MDP is the holding period, which might consist of several trading days. On the other hand, the time horizon for the low-level MDP is a small trading window, e.g., a trading day. The timestep of the low-level MDP could be several minutes for the low-level policy to take actions. We use $t$ as the high-level timestep, and $t'$ as the low-level timestep.

% Note that although many existing works have discussed MDP solution to the portfolio management problem, it is still challenging due to the real-world trading scenarios. First and foremost, each trade cannot be carried out immediately at the last price when a order is placed in the real-world market. Moreover, the existence of transaction cost might turn action changing too much from previous weight vector into sub-optimal action if the transaction cost overweights the immediate return. 

\subsection{The High-level MDP for Portfolio Management}

The high-level MDP is modeled for the portfolio management problem. That is, the trading agent gives new portfolio weights for reallocating the fund into a number of financial assets at the beginning of each holding period.

\textbf{State.} We describe the state $s^h_t = \{\boldsymbol{X}_t, \boldsymbol{w_t}\}\in \mathcal{S}^h$ with historical stock basic features $X_t$, and the current portfolio weights $\boldsymbol{w_t}$. For the $M$ non-cash assets, the features $\boldsymbol{X}_{i,t}$ for asset $i$ are built by a time window $k$: $\boldsymbol{X}_{i,t}=\{\boldsymbol{x}_{i,t-k+1}, \boldsymbol{x}_{i,t-k+2}, \cdots, \boldsymbol{x}_{i,t} \}$, where $\boldsymbol{x}_{i,t}$ presents the basic information of asset $i$ at trading day $t$, including opening, closing, highest, lowest prices and volume. Note that the state can be enriched by adding more factors for better performance. We only use basic daily information to describe high-level state in this work for simplification and fair comparison.

\textbf{Action.} At the end of the holding period $t$, the agent will give a high-level action $a^h_t$ based on the state $s^h_t$ to redistribute the wealth among the assets as a subtask. The subtask is determined by the difference between portfolio weights $\boldsymbol{w'_{t}}$ and $\boldsymbol{w_{t+1}}$. Since $\boldsymbol{w'_{t}}$ has already been determined in the holding period $t$ according to Eq. (\ref{eq:value transistion}), the action of the agent in holding period $t$ can be represented solely by the portfolio vector $\boldsymbol{w_{t+1}}$. Consequently, we define the action as $a^h_t = \boldsymbol{w_{t+1}}$ in the continuous action space, and the action $a_t$ holds the properties in Eq. (\ref{eq:portfolio weights}).

\textbf{Reward.} For each state-action pair, i.e., $(s^h_t, a^h_t)$ at holding period $t$, according to Eq. (\ref{eq:with transaction cost}), $r_t = {v}_{t+1}- {v}_{t}$.

% \begin{figure}[t]
%   \centering
%     % {
%     %   \includegraphics[width=0.47\textwidth]{backtest_us.pdf}
%     % }
%     % \caption{The portfolio value in U.S. market.}
%     % \label{fig:pvus}
%     {
%       \includegraphics[width=0.47\textwidth]{rl_trading.pdf}
%     }
%     % \includegraphics[width=\textwidth]{rl_trading.pdf}  
%     \caption{\textit{HRPM} Mode}
%     \label{fig:hrpm}
% \end{figure}

% \begin{wrapfigure}{r}{0.12\textwidth}
%     \centering
%     \includegraphics[width=\linewidth]{limit_order_book.pdf}  
%     \caption{The limit order book}
%     \label{fig:LOBs}
% \end{wrapfigure}

\subsection{The Low-level MDP for Trade Execution}

The high-level policy assigns subtasks for the low-level policy. The subtasks are represented by an allowed time window with length of $T_{window}$ and the target trading quantity $q_{target}$. $T_{window}$ is based on manual configurations, and $q_{target, t}$ of trading period $t$ can be computed by $q_{target,t} = v'_t |w_{t+1} - w'_{t}| \odot \frac{1}{{p}'_t}$, with assumption of a continuous market. The subtasks can be considered as the trade execution, where the trading agent places many small-sized limit orders at different times at corresponding desired prices. As a result, we model the subtask as the low-level MDP.

% In the subtasks, the low-level policy buys and sells corresponding shares in the market within a fixed time period to minimize the trading part of the whole trading cost. 

% Trade execution at most exchange markets is through limit order books (LOBs), which is a set of all active bid (buy) and ask (sell) orders. A limit order can be modeled as $(p_{lob}, v_{lob}, side_{lob})$, where $p_{lob}$ represents the submitted price, $v_{lob}$ represents the submitted volume, and $side_{lob}$ represents the side of the order. Orders submitted at time $t_{lob}$ are sorted into different levels based on their prices. For instance, the lowest ask price and the highest bid price are grouped into the first level order, followed by the second lowest ask price and the second-highest bid price as the second level, and so on. Consequently, the timeseries evolution of an LOB can be modeled as a 3-dimensional tensor: the first dimension represents time, the second dimension is level, and the third represents prices and order volume on both the buy and sell sides.

\textbf{State.} We maintain a state $s^l_{t'}$ at each low-level time step $t'$, considering the private states and the market states. The private states of trade execution are remaining trading time $T_{window}-T_{elapsed}$ and the remaining quantity $q_{target, t}-q_{finished}$, where $T_{elapsed}$ is the used time and $q_{finished}$ is the finished quantity. The market states consist of the historical limit order books and multiple technical factors as shown in Fig. \ref{fig:LOBs}. Specifically, we collect historical LOBs in a time window $k^{\prime}$: $\boldsymbol{LOB}_{t'}=\{\boldsymbol{lob}_{t'-k}, \boldsymbol{lob}_{t'-k+1}, \cdots, \boldsymbol{lob}_{t'-1} \}$, where $\boldsymbol{lob}_{t'}$ is the limit order book at time step $t'$, including price and volume.

% For instance, the lowest ask price and the highest bid price are grouped into the first level order, followed by the second lowest ask price and the second-highest bid price as the second level, and so on. Consequently, the timeseries evolution of an LOB can be modeled as a 3-dimensional tensor: the first dimension represents time, the second dimension is level, and the third represents prices and order volume on both the buy and sell sides.

% here limit order figure

\textbf{Action.} Each low-level action corresponds to a limit order decision with the target price $p_{target}$ and the target quantity $\pm q_{target}$. Note that a zero quantity indicates that we skip the current trading time step with no order placement, and if an action $a_{t'}$ fails to be executed in the low-level time step $t'$ due to an inappropriate price, the action will expire at the next low-level time step $(t'+1)$ without actual trading. Any quantity remaining at the end of the trading period must be cleaned up by using a market order, walking through the lower prices on the bid side of the order book until all remaining volumes are sold.

\textbf{Reward.} Once receiving a limit order decision, the environment will match this order and feedback an executed price $p_{paid}$. We define the low-level reward as: $r^l_{t'} = -\left[\lambda(\pm q_{target} \times p_{paid})+ (p_{paid} - p_{t+1}) \times (\pm q_{target})\right] $. So the cumulative trading cost in a low-level episode will be reported to the high-level policy to compute the high-level reward $c_{trade, t}=-\sum_{t'}^{T^l}r^l_{t'}$.

% From a very high level perspective, it is obvious that for every order there is an optimal execution rate or execution schedule, that is, speed with which order is executed, or the duration of its execution in the marketplace.

% Once a rough optimal rate or schedule is found, the next level of decision-making deals with the implementation of the schedule. In order to stay on schedule, the agent typically tries to blend with the rest of the market: being an outlier is penalized because it reveals the agent’s intention. The agent creates marketplace orders which mimic other participants’ orders — both in size and in prices.

\section{Optimizing via Deep Hierarchical Reinforcement Learning}

In this section, we first present our solution for learning hierarchical policies, then propose our solution algorithms respectively for the two levels of MDPs. Finally, we introduce two schemes: pre-training and iterative training for better financial data usage.

\subsection{Hierarchy of Two Policies}
Our environment has two dynamics: the price dynamic and the limit order book dynamic. We extend the standard RL setup to a hierarchical two-layer structure, with a high-level policy $\pi^{h}$ and a low-level policy $\pi^{l}$. Different from the existing HRL algorithms, our high-level policy and the low-level policy run in different timescales. 
% Our environment has two dynamics: the price dynamic and the limit order book dynamic. 
% The high-level policy operates at a low frequency and sets goals for the low-level policy. The goals correspond directly to private states that the low-level policy attempts to fulfill. 
As shown in Fig. \ref{fig:hrpm}, each high-level time step $t$ consists of three steps:

\begin{enumerate}
    \item The higher-level policy observes the high-level state $s^h_{t}$ and produces a high-level action $a^h_t$, which corresponds to the new portfolio for the next step. 
    \item The high-level action $a^h_t$ would generate subtasks for the low-level policy according to the gap between the current portfolio and the new one.
    \item The low-level policy produces the low-level action $a^l_{t'}$ based on the low-level states $s^l_{t'}$. These actions would be applied to the limit order dynamic for updating the current portfolio. 
\end{enumerate}

\subsection{High-level RL with Entropy Bonus}

We consider the high-level problem as a continuous control problem, since the high-level policy is supposed to generate a continuous vector according to Eq. (\ref{eq:portfolio weights}). Previous works utilize DDPG \cite{lillicrap2015continuous} to generate portfolio weights. Unfortunately, DDPG faces challenges in solving our problem. First, DDPG highly relies on a well-trained critic. However, the critic also needs lots of data, which is limited in the financial area \cite{liang2018adversarial}. Additionally, the exploration in DDPG is dependant on the Gaussian noise when applying actions into the environment. This exploration might fail due to choosing specific sequences of portfolio weights and excluding other possible sequences of portfolio weights. Also, we find that the portfolio weights will converge to a few assets. That is, the agent only holds few assets. 

Here we utilized the REINFORCE algorithm \cite{sutton2000policy}. One approach to address these issues is to introduce an entropy bonus. In order to encourage the high-level policy not to ``put all the eggs in one basket", we aim to find a high-level policy that maximizes the maximum entropy objective:

\begin{small}
\begin{equation}
\begin{aligned}
{\pi^{h}}^{*}&=\underset{\pi^h}{\operatorname{argmax}} \sum_{t=0}^{T^h} E_{s^h_{0} \sim \beta_{0}, s^h_{t+1} \sim p^h, a^h_{t} \sim \pi^h}[\gamma^{h}\tilde{r}^h (s^h_{t}, a^h_{t})] \\
&=\underset{\pi^h}{\operatorname{argmax}} \sum_{t=0}^{T^h} E_{s^h_{0}, s^h_{t+1}, a^h_{t}}\left[\gamma^{h}\left(r^h(s^h_{t}, a^h_{t})+\eta \mathcal{H}(a^h_{t})\right)\right]
\end{aligned}
\end{equation}
\end{small}\normalsize
where $\beta_0$ is the intial high-level state distribution, $s^h_{t+1} \sim p^h$ is an abbreviation of $s^h_{t+1} \sim p^h(\cdot|s^h_{t}, a^h_{t})$, $a^h_{t} \sim \pi^h$ is an abbreviation of $a^h_{t} \sim \pi^h(\cdot|s^h_t)$, $\eta$ determines the relative importance of the entropy term versus the reward, $\mathcal{H}(a^h_t)$ is the entropy of the portfolio weights, which is calculated as $\mathcal{H}(a^h_t) = \mathcal{H}(\boldsymbol{w^h_{t+1}}) = -\sum_{i} w^h_{i, t+1} \log w^h_{i, t+1}$, and $\tilde{r}^h$ is an abbreviation of the summation of the high-level reward and entropy bonus.

We use the softmax policy as the high-level policy $\pi^h(s^h, a^h; \theta)=\frac{e^{\theta^{T} \cdot \phi(s^h, a^h)}}{\sum_{b} e^{\theta^{T} \cdot \phi(s^h, b)}} \text { for all } s^h \in \mathcal{S}^h, a^h \in \mathcal{A}^h$. Let $J(\pi^h_{\theta})=\mathbb{E}_{\pi^h}\left[\sum_{t=0}^{T} \gamma^{t} t_{t}\right]$ denote the expected finite-horizon discounted return of the policy. According to the policy gradient theorem \cite{sutton2000policy}, we compute the gradient $\nabla_{\theta} J\left(\pi^h_{\theta}\right)=\underset{\tau \sim \pi_{\theta}}{\mathrm{E}}\left[\sum_{t=0}^{T} \nabla_{\theta} \log \pi^h_{\theta}\left(a_{t} | s_{t}\right) G_t\right]$, where $G_t=\sum_{j=0}^{T-t}\gamma^j\tilde{r}^h$. Thus we update the high-level policy parameters by $\theta \leftarrow \theta+\alpha \gamma^{t} \nabla_{\theta} \log \pi^h_{\theta}\left(a_{t} | s_{t}\right) G_t$.

\subsection{Low-level RL with Action Branching}
The low-level problem is considered as a discrete control problem with two action dimensions: price and quantity. We utilize the Branching Dueling Q-Network \cite{tavakoli2018action}. Formally, we have two action dimensions with $|p^l| = n_p$ discrete relative price levels and $|q^l| = n_q$ discrete quantity proportions. The action value $Q^l_d$ at state $s^l \in \mathcal{S}^l$ and the action $a^l_{d} \in \mathcal{A}^l_{d}$ are expressed in terms of the common state value $V^l(s)$ and the corresponding (state-dependent) action advantage $Adv^l_{d}\left(s, a^l_{d}\right)$ for $d \in \{p,q\}$:

\begin{small}
\begin{equation}\nonumber
Q^l_{d}(s^l, {a^l_{d}})=V(s^l)+(Adv_{d}(s^l, {a^l_{d}})-\frac{1}{n} \sum_{{a^l_{d}}^{\prime} \in \mathcal{A}_{d}} Adv_{d}(s^l, {a^l_{d}}^{\prime}))
\end{equation}
\end{small}\normalsize
We train our Q-value function based on the one-step temporal-difference learning:
\begin{small}
\begin{equation}
y_{d}=r+\gamma Q_{d}^{-}({s^l}^{\prime}, \underset{{a^l_{d}}^{\prime} \in \mathcal{A}_{d}}{\arg \max } Q_{d}({s^l}^{\prime}, {a^l_{d}}^{\prime})), d\in\{p,q\}
\end{equation}
\end{small}\normalsize
\begin{small}
\begin{equation}
L=\mathbb{E}_{(s, a, r, s^{\prime}) \sim \mathcal{D}}[\frac{1}{N} \sum_{d\in\{P,I\}}(y_{d}-Q_{d}({s^l}, a^l_{d}))^{2}]
\end{equation}
\end{small}\normalsize
where $\mathcal{D}$ denotes a (prioritized) experience replay buffer and $a$ denotes the joint-action tuple $\left(p^l,q^l\right)$.

\subsection{Training Scheme}

The financial data, including price information, multiple factors, and limit order book, is complex and high-dimensional. In order to use the available data more efficiently, we utilize the pre-training scheme and the iterative training. 

\noindent \textbf{Pre-training.}  In this work, we pre-train the low-level policy in the limit order book environment. Specifically, we first pre-select several assets, i.e., stocks, as our pre-trained data sources. By individually training the corresponding buying/selling policy of each asset, we derive multiple low-level policy parameters for these assets. In the general setting of hierarchical reinforcement learning, the high-level policy and the low-level policy are only trained together in a single environment. However, this training scheme suffers from data insufficiency, since the agent's interactions with the environment are severely restricted due to the joint training, especially for the low-level policy training. By introducing the pre-training scheme, the low-level policy would be trained with more and diverse interactions, thereby having better generalization and robustness.

\noindent \textbf{Iterative training.} Following the iterative training scheme in \cite{nevmyvaka2006reinforcement}, we augment the private state repeatedly in the low-level pre-training. Specifically, we traverse the target quantity from $(0,0)$ to a max target quantity $q_{max}$ and the remaining time from 0 to a max trading time window $T'_{max}$. In this way, the low-level policy is trained with the augmented private states, thus can generalize different subtasks assigned by the high-level policy.

% We notice that in end-to-end training, the low-level policy can only explore rare private states. Thus, based on the pre-training scheme, we follow the iterative training scheme in \cite{nevmyvaka2006reinforcement} by repeatedly augmenting the private state in the low-level pre-training. We first pre-select a max target inventory $I_{max}$ and a max tarding timewindow $T'_{max}$. Then we slice $I_{max}$ and $T'_{max}$ into many pieces with the equal lengths $i_{interval}$ and $t'_{interval}$. The low-level policy is trained by starting from $s^l_{private} = (i_{interval}, t'_{interval})$ to $s^l_{private} = (I_{max}, T'_{max})$. It means that the low-level policy is trained with an augmented private state. Hence, the low-level has learned each discrete step of inventory and timewindow in the process of buying or selling $I_{max}$ within $T'_{max}$.

% Given that our trading system is developed in different levels of granularity, 

% Training an RL agent required a huge number of interactions with the environment. Moreover, 

% Current RL algorithms are often limited by its data inefficiency, i.e., the required number of interactions with
% the environment is impractically high. Moreover, 

% Training an RL agent requires a number of episodic rollouts each of which cannot be parallelized due to the feedback loop between the agent and its environment.

% RL needs data. Finance needs interpretability. Current HRL end-to-end, one to many, however, in our task, one to one. Also low-level only train with latent goal, break interpretability.

\section{Experiment}

In this section, we introduce our data processing, baselines, and metrics. Then we evaluate our algorithm with baselines in different markets. Lastly, we analyze the impact of entropy in reward and the trading cost by ablation study.

\subsection{Dataset setting and Preprosessing}

Our experiments are conducted on stock data from the U.S. stock market and China stock market. The stock data are collected from wind\footnote{https://www.wind.com.cn/}. We evaluate our algorithm on different stock markets to demonstrate robustness and practicality. We choose stocks with large volumes so that our trading action would not affect the market price. Specifically, we choose 23 stocks from \textit{Dow Jones Industrial Average Index (DJIA)} and 23 stocks from \textit{SSE 50 Index} for the U.S. stock market and China stock market respectively. In addition, we introduce 1 cash asset as a risk-free choice for both of the two markets. Moreover, the period of stock data used in the experiments is shown in Table \ref{tab:period}.

\begin{table}[ht]
\small
\centering

\begin{tabular}{c|c|c}
\hline
\hline
 & \textsl{the U.S. Market} & \textsl{the China Market} \\
\hline
Training & 2000/01/03-2014/01/03 & 2007/02/08-2015/07/03 \\
\hline
Evaluate &  2014/01/06-2015/12/31 & 2015/07/06-2017/07/03 \\
\hline
Test &  2016/01/04-2018/05/22 & 2017/07/04-2018/07/04 \\
\hline
\hline
\end{tabular}

\caption{Period of stock data used in the experiments.}\label{tab:period}
\end{table}

For the high-level setting, we set the holding period to 5 days and set the trading period to one day. We set the time window of the high-level state as 10 days. We normalize the stock data to derive a general agent for different stocks in data pre-processing. Specifically, we divide the opening price, closing price, high price, and low price by the close price on the first day of the period to avoid the leakage of future information. Similarly, we use the volume on the first day to normalize the volume. For missing data which occurs during weekends and holidays, we fill the empty price data with the close price on the previous day and set zero volume to maintain the consistency of time series . For the constant commission rate, we set it to $0.2\%$ on both two markets.

For the low-level setting, we set the time interval to 30 seconds. We set the time window of the low-level state as opening hours of one day. Similarly, we also normalize the prices in limit order books by dividing the first price at the beginning of the period, as well as the volume. For missing data which occurs during weekends and holidays, we fill the empty price data with the previous price and set zero volume to maintain the consistency of time series.

\subsection{Baseline and Metrics}

\noindent \textbf{Baselines.} We compare our methods with the baselines:

\begin{itemize}
    \item \textit{Uniform Constant Rebalanced Portfolios (UCRP)} \cite{cover2011universal} keeps the same distribution of wealth among a set of assets from day to day.
    
    \item \textit{On-Line Moving Average Reversion (OLMAR)} \cite{li2012line} exploits moving average reversion to overcome the limitation of single period mean reversion assumption.
    
    \item \textit{Weighted Moving Average Mean Reversion (WMAMR)} \cite{gao2013weighted} exploits historical price information by using equal weighted moving averages and then learns portfolios by online learning techniques.
    
    \item \textit{Follow the Winner (Winner)} \cite{gaivoronski2000stochastic} always focus on the outperforming asset and transfers all portfolio weights to it.
    
    \item \textit{Follow the Loser (Loser)} \cite{borodin2004can} moves portfolio weights from the outperforming assets to the underperforming assets.

    \item \textit{Deep Portfolio Management(DPM)} \cite{jiang2017deep} is based on Ensemble of Identical Independent Evaluators (EIIE) topology. \textit{DPM} uses asset prices as state and trains an agent with a Deep Neural Network (DNN) approximated policy function to evaluate each asset’s potential growth in the immediate future and it is the state-of-the-art RL algorithm for PM.
\end{itemize}

In addition, \textit{SSE 50 Index} and \textit{Dow Jones Industrial Average Index (DJIA)} are introduced as baselines in the comparison of performance to demonstrate whether these portfolio strategies are able to outperform the market.

\begin{figure}[t]
  \centering
    {
       \includegraphics[width=0.47\textwidth]{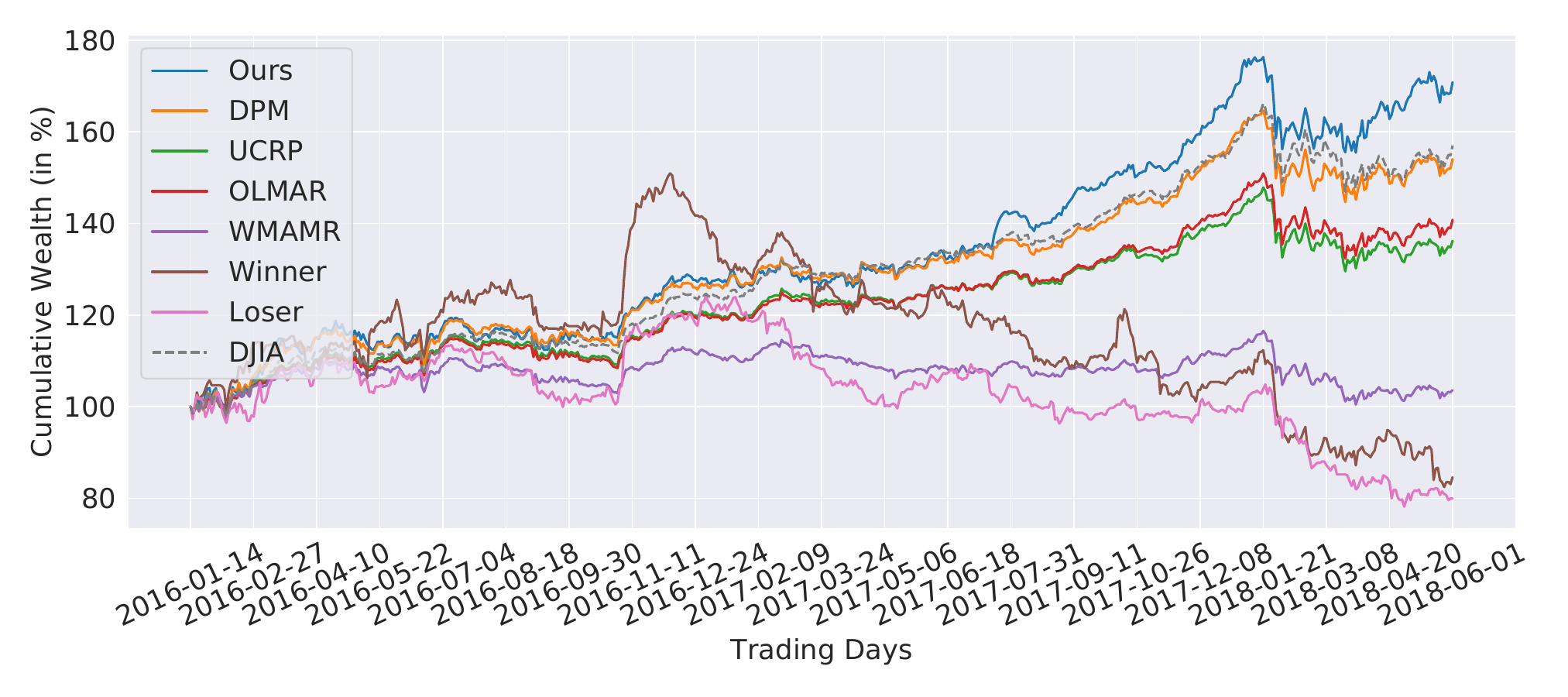}
    }
    \caption{The portfolio value in the U.S. market.}
    \label{fig:pvus}
\end{figure}

\begin{table}[t]
\centering
\scalebox{0.9}{
\begin{tabular}{cccccc}
\hline
\hline
    & ARR(\%) & ASR & MDD & DDR       \\ \hline
UCRP & 13.827 &  1.24  & 0.124 & 1.348 \\
Winner & -6.773 & -0.474 & 0.454  & -0.455    \\
Loser & -8.961 & -0.911  & 0.371 & -0.723  \\
OLMAR & 15.428 & 1.252 & 0.123  & 1.503    \\
WMAMR & 1.503 & 0.468  & 0.138 & 0.143  \\
DPM & 19.864 & \textbf{1.263}  & 0.121 & 1.815 \\
\textbf{HRPM} & \textbf{27.089} & {1.246}  & \textbf{0.117} & \textbf{2.403}  \\
\hline
\textit{DJIA} & \textit{20.823} &  \textit{1.206}  & \textit{0.116} & \textit{2.547} \\
\hline
\hline
\end{tabular}
}
\caption{Performance comparison in the U.S. market}\label{tab:uspc}
\end{table}

\noindent \textbf{Metrics.} we use the following performance metrics for evaluations: \textit{Annual Rate of Return (ARR)}, \textit{Annualized Sharpe Ratio (ASR)}, \textit{Maximum DrawDown (MDD)} and \textit{Downside Deviation Ratio (DDR)}.

\begin{itemize}
    \item \textit{Annual Rate of Return (ARR)} is an annualized average of return rate. It is defined as ${ARR} = \frac{V_f-V_i}{V_i} \times \frac{T_{year}}{T_{all}}$, where $V_f$ is final portfolio value, $V_i$ is initial portfolio value, $T_{all}$ is the total number of trading days, $T_{year}$ is the number of trading days in one year.
    \item \textit{Annualized Sharpe Ratio (ASR)} is defined as \textit{Annual Rate of Return} divided by the standard deviation of the investment. It represents the additional amount of return that an investor receives per unit of increase in risk.
    \item \textit{Maximum DrawDown (MDD)} is the maximum observed loss from a peak to a bottom of a portfolio, before a new peak is attained. It is an indicator of downside risk over a specified time period.
    \item \textit{Downside Deviation Ratio (DDR)} measures the downside risk of a strategy as the average of returns when it falls below a \textit{minimum acceptable return (MAR)}. It is the risk-adjusted ARR based on Downside Deviation. In our experiment, the \textit{MAR} is set to zero.
\end{itemize}

% We test both baseline models and our methods in 2019, and we calculate the trade returns net of transaction cost. We then form a simple portfolio by giving equal weights to each contract, and the trade return of a portfolio is:

\subsection{Comparison with Baselines}

\noindent \textbf{Backtest Results in the U.S. Market.} Figure \ref{fig:pvus} shows the cumulative wealth vs. trading days in the U.S. market. From the plots of \textit{DJIA} index, we can see this period is in a bull market generally, although the market edges down several times. While simple methods like \textit{Following the Winner} and \textit{Following the Loser} fail, the portfolio values conducted by most of the strategies keep climbing but still underperform the market. Particularly, our \textit{HRPM} is the only strategy that outperforms the \textit{DJIA} index, when \textit{DPM} keeps almost the same as the trend of the market.

We can compare the metrics of different strategies in detail in Table \ref{tab:uspc}. On \textit{ARR}, \textit{HRPM} performs much better than the other methods and outperforms the market with 6.2\%. On \textit{ASR}, most of methods get higher scores than \textit{DJIA} index and \textit{HRPM} is the best. That is, our strategy could gain more profit under the same risk. When it goes to \textit{MDD} and \textit{DDR}, the results show that \textit{HRPM} bears the least risk, even lower than \textit{UCRP}. The risk of our method is acceptable, although a little higher than the \textit{DJIA} index.

\begin{figure}[t]
  \centering
    {
       \includegraphics[width=0.47\textwidth]{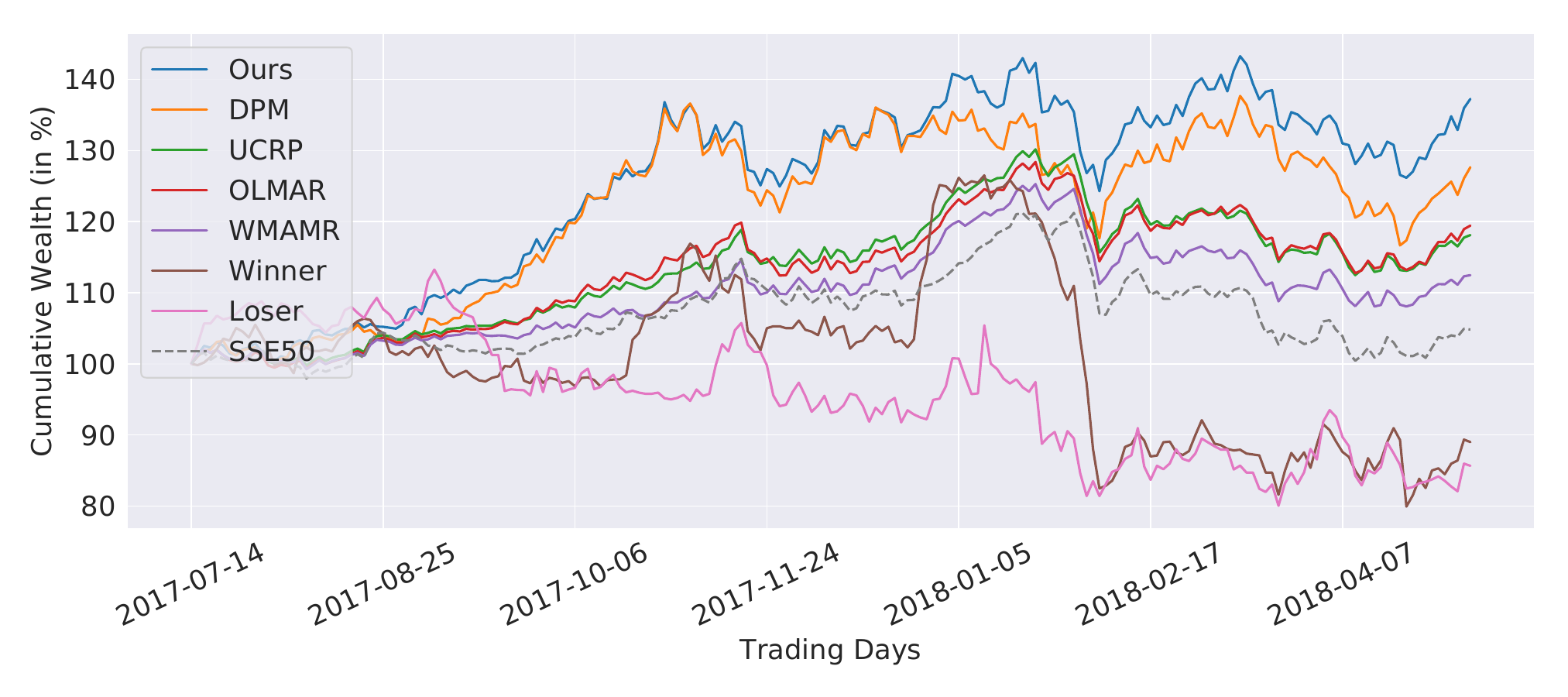}
    }
    \caption{The portfolio value in the China market.}
    \label{fig:pvchina}
\end{figure}

\begin{table}[t]
\centering
\scalebox{0.9}{
\begin{tabular}{cccccc}
\hline
\hline
    & ARR(\%) & ASR & MDD & DDR       \\ \hline
UCRP & 23.202 &  2.917  & 0.136 & 1.863 \\
Winner & -13.6 & -1.201 & 0.368  & -0.705    \\
Loser & -17.645 & -2.1  & 0.293 & -1.045  \\
OLMAR & 24.947 & 3.252 & 0.122  & 1.961    \\
WMAMR & 15.873 & 2.485  & 0.137 & 2.775  \\
DPM & 35.81 & 3.126  & 0.153 & 2.404 \\
\textbf{HRPM} & \textbf{48.742} & \textbf{3.813}  & \textbf{0.131} & \textbf{3.325}  \\
\hline
\textit{SSE50} & \textit{6.081} &  \textit{1.1}  & \textit{0.172} & \textit{0.641} \\
\hline
\hline
\end{tabular}
}
\caption{Performance comparison in the China market}\label{tab:chinapc}
\end{table}

\noindent \textbf{Backtest Results in the China Market. } Figure \ref{fig:pvchina} shows cumulative wealth vs. trading days in the China market. As we can see, \textit{SSE50} index first rises slowly, but from then on it falls back to the starting point. During the rising stage, \textit{HRPM} keeps the best performance and the longest time. When the portfolio values of other methods decline, our strategy still hovers at the peak. This phenomenon demonstrates that \textit{HRPM} is not only superior to all the baselines, it is also robust under different market conditions relatively.

Table \ref{tab:chinapc} compares all metrics of different strategies in detail. On all the four metrics, our \textit{HRPM} achieves the best among all the strategies. Specifically, \textit{HRPM} outperforms \textit{DPM} by $13\%$ and outperforms market by $42\%$ on \textit{ARR}. Even considering return with risk by the \textit{ASR} metric, our strategies work better than others. Different from the performance in the U.S. market, \textit{HRPM} not only reaches the least risk among these strategies but also performs better than the market on \textit{MDD} and \textit{DDR}. It means that our strategy is able to adapt to volatile markets with low risk.

\begin{figure}[!t]
  \centering
    {
       \includegraphics[width=0.47\textwidth]{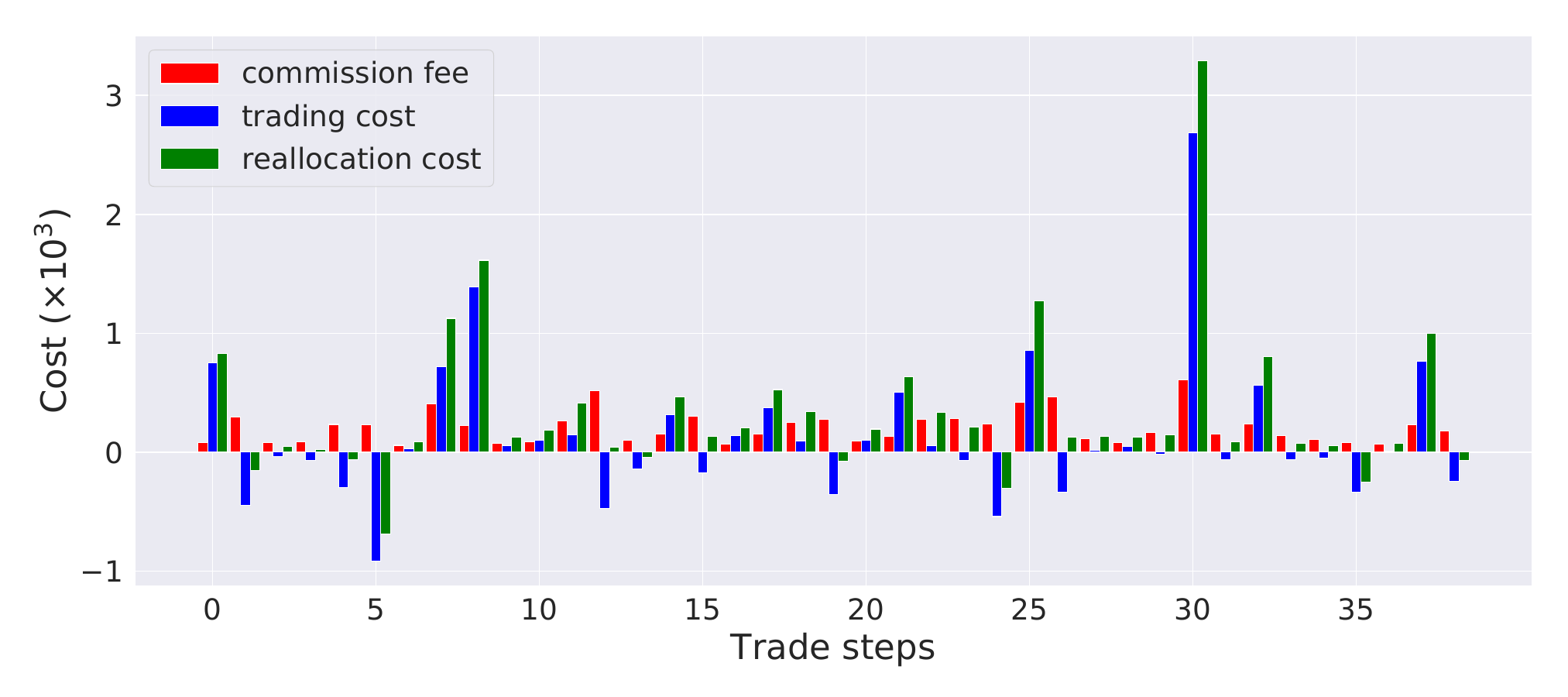}
    }
    \caption{Trading cost during the trading periods.}
    \label{fig:tc}
\end{figure}

\begin{figure}[!t]
  \centering
    {
       \includegraphics[width=0.47\textwidth]{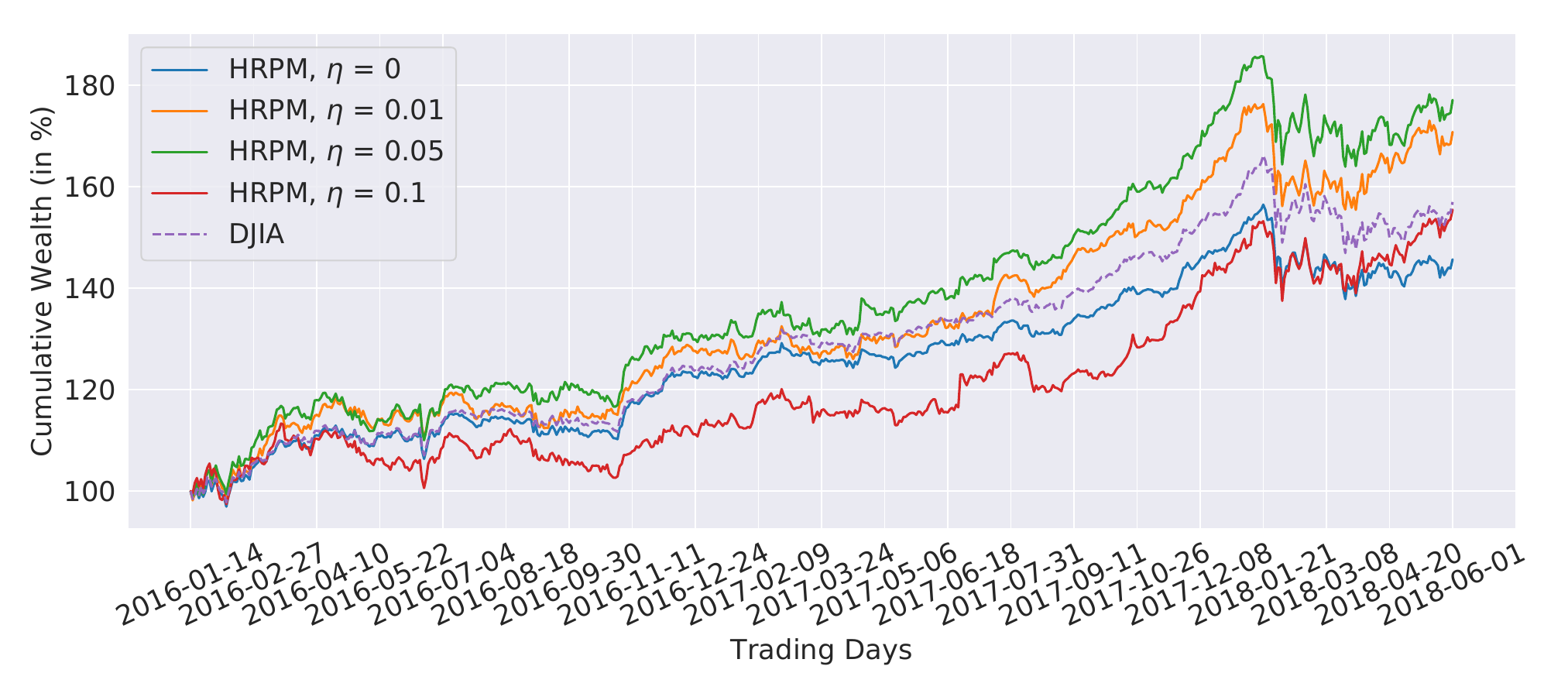}
    }
    \caption{The effect of the entropy bonus in the U.S. Market.}
    \label{fig:entropy}
\end{figure}

\begin{table}[!t]
\centering
% \scalebox{0.9}{
\begin{tabular}{cccccc}
\hline
\hline
    & ARR(\%) & ASR & MDD & DDR       \\ \hline
HRPM, $\eta$ = 0 & 17.092 &  1.207  & 0.191 & 1.631 \\
HRPM, $\eta$ = 0.01  & 25.172 & 1.201 & 0.183  & 2.275   \\
HRPM, $\eta$ = 0.05 & 27.089 & 1.246  & 0.117 & 2.403  \\
HRPM, $\eta$ = 0.1 & 20.33 & 1.346 & 0.124  & 1.754   \\
\hline
\textit{DJIA} & \textit{20.823} &  \textit{1.206}  & \textit{0.116} & \textit{2.547} \\
\hline
\hline
\end{tabular}
% }
\caption{Ablation on the effect of entropy in the U.S. market}\label{tab:uspc}
\label{tb:entropy}
\end{table}

\subsection{Ablation Study}

In our ablation study, we mainly answer two questions: (1) why the commission fee is not enough to account for the actual trading cost? (2) Does the entropy bonus work?

\noindent \textbf{Commission Fee is Not Enough.} In Fig. \ref{fig:tc}, we show the trading cost at each trading period of testing in the China market. The x-axis is the index of each trading period in testing, and the y-axis is the cost. Note that a negative trading cost means gaining profit in the trading period. A negative trading cost happens when the submitted orders are finished at a better price than the target price. We can see that the commission fee is only a small part of the total trading cost in most transactions, especially in large transactions.

% An interesting phenomenon is that the high-level agent tends to give a large reallocation task when market is in a peak or valley, e.g., the agent makes the largest reallocation in the 30-th trade step, which corresponding to \textit{2008-02-06} in Figure \ref{fig:pvchina}.

% Only considering the commission fee is not an accurate approximation to the actual transaction cost. 

\noindent \textbf{Entropy Bonus.} We show the effect of the entropy bonus in Fig. \ref{fig:entropy}. As $\eta$ in the entropy bonus term in reward gets higher, the agent would tend to give portfolios with more diversity. While $\eta$ is small, the agent might be more likely to hold only a few stocks. Consequently, we may get a balance between the risk-control and profit by controlling $\eta$. As Table \ref{tb:entropy} shows, \textit{HRPM} with $\eta=0.05$ reaches the maximum ARR. Note that \textit{HRPM} with $\eta=0.1$ has the lowest MDD than others. By increasing the effect of the entropy, we can gain more profit relative to the risk.

\begin{figure}[!htb]
  \centering
     \subfloat[{Selling 40,000 shares}\label{fig:antmaze2}]{%
      \includegraphics[width=0.22\textwidth]{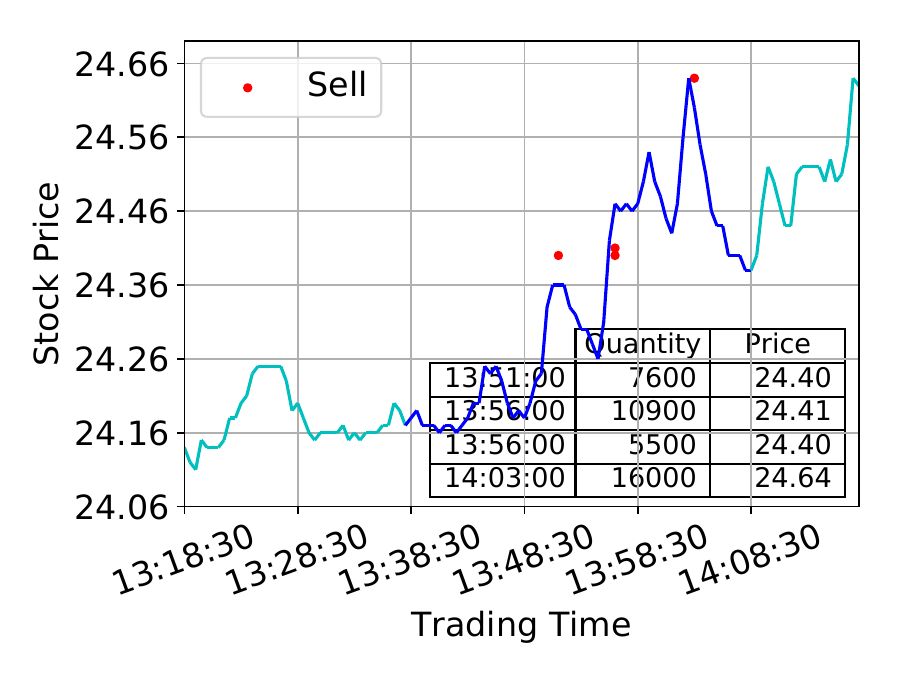}
     }
     \hfill
     \subfloat[{Buying 40,000 shares}\label{fig:antmaze1}]{%
      \includegraphics[width=0.22\textwidth]{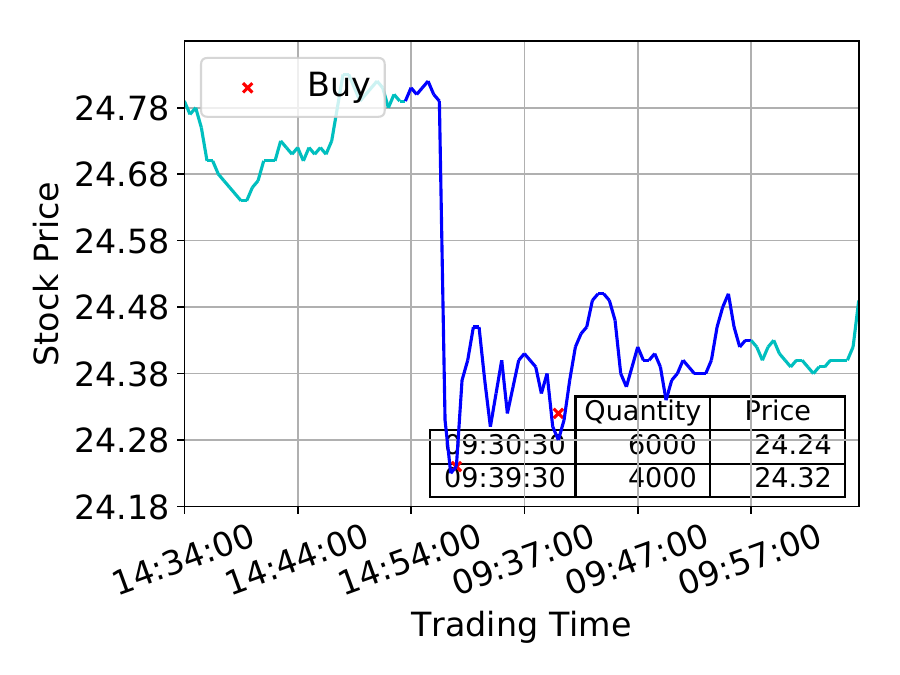}
     }
     \hfill
     \subfloat[{Selling 10,000 shares}\label{fig:antmaze0}]{%
      \includegraphics[width=0.22\textwidth]{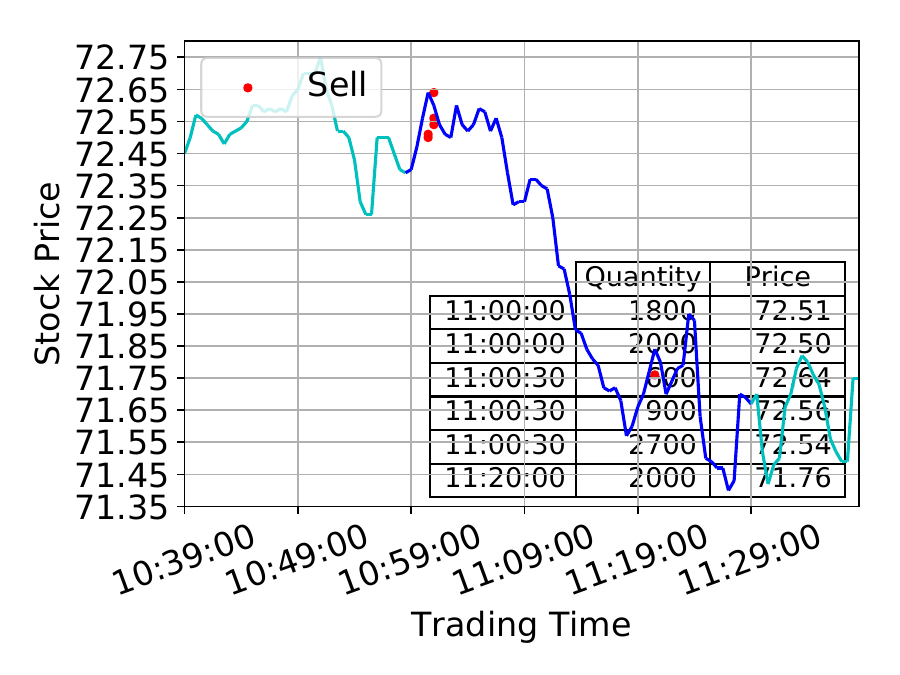}
     }
     \hfill
     \subfloat[{Buying 10,000 shares}\label{subfig-1:dummy}]{%
      \includegraphics[width=0.22\textwidth]{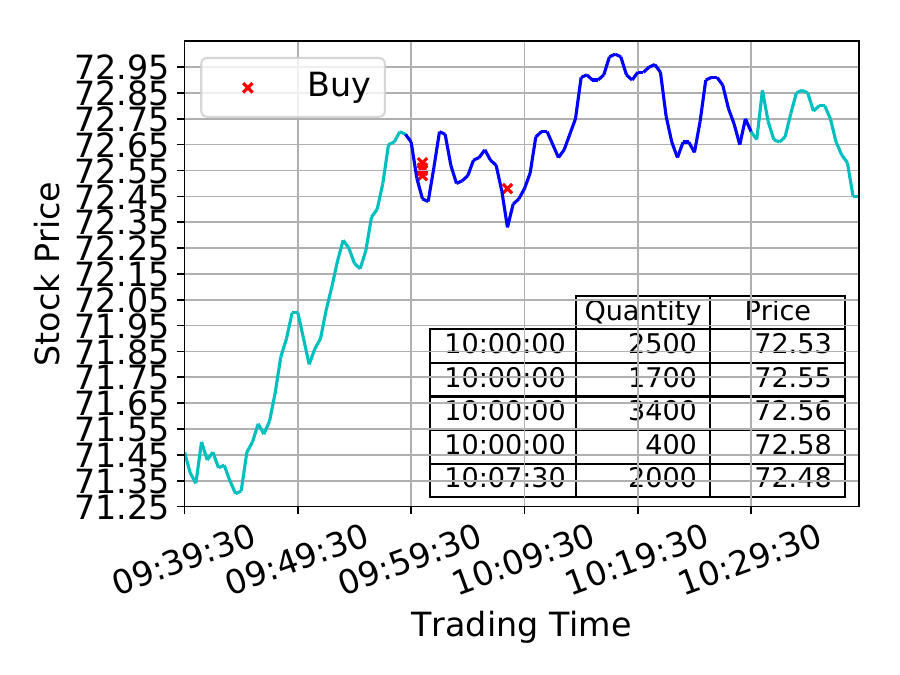}
     }
     \caption{\small{The interpretation of the low-level policy. The line is the mid-price of the market. The dark blue part is the trading window, and the light blue part shows the price before and after. The table inside the figure shows the finished trades.}}
     \label{fig:low-level strategy}
\end{figure}

\subsection{Trading Strategy Interpretation}

Here, we try to interpret the underlying investment strategies of the low-level policy. We show the trades that happened in the trading period in Fig. \ref{fig:low-level strategy}. Fig. \ref{fig:low-level strategy}(a)(c) shows that the low-level policy tries to sell orders at a relatively high position. It first sells part of the quantity at the beginning of the trading period. Then it sells the rest at the following local maximum. In this way, the low-level policy can avoid missing potential growth and decline. We can get the same conclusion for buying in Fig. \ref{fig:low-level strategy}(b)(d).

\section{Conclusion}

In this paper, we focus on the problem of portfolio management with trading cost via deep reinforcement learning. We propose a hierarchical reinforced stock trading system (\textit{HRPM}). Concretely, we build a hierarchy of portfolio management over trade execution and train the corresponding policies. The high-level policy gives portfolio weights and invokes the low-level policy to sell or buy the corresponding shares within a short time window. Extensive experimental results in the U.S. market and China market demonstrate that \textit{HRPM} achieves significant improvement against many state-of-the-art approaches. 

% In future work, we will explore the jointly training scheme and incorporate multi-factor models into the hierarchical framework.

\section*{Acknowledgements}

This research is supported, in part, by the Joint NTU-WeBank Research Centre on Fintech (Award No: NWJ-2019-008), Nanyang Technological University, Singapore.

\bibliography{reference.bib}

\begin{thebibliography}{20}
\providecommand{\natexlab}[1]{#1}
\providecommand{\url}[1]{\texttt{#1}}
\providecommand{\urlprefix}{URL }
\expandafter\ifx\csname urlstyle\endcsname\relax
  \providecommand{\doi}[1]{doi:\discretionary{}{}{}#1}\else
  \providecommand{\doi}{doi:\discretionary{}{}{}\begingroup
  \urlstyle{rm}\Url}\fi

\bibitem[{Almahdi and Yang(2017)}]{almahdi2017adaptive}
Almahdi, S.; and Yang, S.~Y. 2017.
\newblock An adaptive portfolio trading system: A risk-return portfolio
  optimization using recurrent reinforcement learning with expected maximum
  drawdown.
\newblock \emph{Expert Systems with Applications} 87: 267--279.

\bibitem[{Borodin, El-Yaniv, and Gogan(2004)}]{borodin2004can}
Borodin, A.; El-Yaniv, R.; and Gogan, V. 2004.
\newblock Can we learn to beat the best stock.
\newblock In \emph{Advances in Neural Information Processing Systems},
  345--352.

\bibitem[{Cover(2011)}]{cover2011universal}
Cover, T.~M. 2011.
\newblock Universal portfolios.
\newblock In \emph{The Kelly Capital Growth Investment Criterion: Theory and
  Practice}, 181--209. World Scientific.

\bibitem[{Gaivoronski and Stella(2000)}]{gaivoronski2000stochastic}
Gaivoronski, A.~A.; and Stella, F. 2000.
\newblock Stochastic nonstationary optimization for finding universal
  portfolios.
\newblock \emph{Annals of Operations Research} 100(1-4): 165--188.

\bibitem[{Gao and Zhang(2013)}]{gao2013weighted}
Gao, L.; and Zhang, W. 2013.
\newblock Weighted moving average passive aggressive algorithm for online
  portfolio selection.
\newblock In \emph{2013 5th International Conference on Intelligent
  Human-Machine Systems and Cybernetics}, volume~1, 327--330. IEEE.

\bibitem[{Jiang and Liang(2017)}]{jiang2017cryptocurrency}
Jiang, Z.; and Liang, J. 2017.
\newblock Cryptocurrency portfolio management with deep reinforcement learning.
\newblock In \emph{2017 Intelligent Systems Conference (IntelliSys)}, 905--913.
  IEEE.

\bibitem[{Jiang, Xu, and Liang(2017)}]{jiang2017deep}
Jiang, Z.; Xu, D.; and Liang, J. 2017.
\newblock A deep reinforcement learning framework for the financial portfolio
  management problem.
\newblock \emph{arXiv preprint arXiv:1706.10059} .

\bibitem[{Li and Hoi(2012)}]{li2012line}
Li, B.; and Hoi, S.~C. 2012.
\newblock On-line portfolio selection with moving average reversion.
\newblock In \emph{Proceedings of the 29th International Coference on
  International Conference on Machine Learning}, 563--570.

\bibitem[{Liang et~al.(2018)Liang, Chen, Zhu, Jiang, and
  Li}]{liang2018adversarial}
Liang, Z.; Chen, H.; Zhu, J.; Jiang, K.; and Li, Y. 2018.
\newblock Adversarial deep reinforcement learning in portfolio management.
\newblock \emph{arXiv preprint arXiv:1808.09940} .

\bibitem[{Lillicrap et~al.(2015)Lillicrap, Hunt, Pritzel, Heess, Erez, Tassa,
  Silver, and Wierstra}]{lillicrap2015continuous}
Lillicrap, T.~P.; Hunt, J.~J.; Pritzel, A.; Heess, N.; Erez, T.; Tassa, Y.;
  Silver, D.; and Wierstra, D. 2015.
\newblock Continuous control with deep reinforcement learning.
\newblock \emph{arXiv preprint arXiv:1509.02971} .

\bibitem[{Mnih et~al.(2013)Mnih, Kavukcuoglu, Silver, Graves, Antonoglou,
  Wierstra, and Riedmiller}]{mnih2013playing}
Mnih, V.; Kavukcuoglu, K.; Silver, D.; Graves, A.; Antonoglou, I.; Wierstra,
  D.; and Riedmiller, M. 2013.
\newblock Playing Atari with deep reinforcement learning.
\newblock \emph{arXiv preprint arXiv:1312.5602} .

\bibitem[{Mosavi et~al.(2020)Mosavi, Ghamisi, Faghan, and
  Duan}]{mosavi2020comprehensive}
Mosavi, A.; Ghamisi, P.; Faghan, Y.; and Duan, P. 2020.
\newblock Comprehensive Review of Deep Reinforcement Learning Methods and
  Applications in Economics.
\newblock \emph{arXiv preprint arXiv:2004.01509} .

\bibitem[{Nevmyvaka, Feng, and Kearns(2006)}]{nevmyvaka2006reinforcement}
Nevmyvaka, Y.; Feng, Y.; and Kearns, M. 2006.
\newblock Reinforcement learning for optimized trade execution.
\newblock In \emph{Proceedings of the 23rd International Conference on Machine
  Learning}, 673--680.

\bibitem[{Silver et~al.(2016)Silver, Huang, Maddison, Guez, Sifre, Van
  Den~Driessche, Schrittwieser, Antonoglou, Panneershelvam, Lanctot
  et~al.}]{silver2016mastering}
Silver, D.; Huang, A.; Maddison, C.~J.; Guez, A.; Sifre, L.; Van Den~Driessche,
  G.; Schrittwieser, J.; Antonoglou, I.; Panneershelvam, V.; Lanctot, M.;
  et~al. 2016.
\newblock Mastering the game of Go with deep neural networks and tree search.
\newblock \emph{Nature} 529(7587): 484.

\bibitem[{Sutton et~al.(2000)Sutton, McAllester, Singh, and
  Mansour}]{sutton2000policy}
Sutton, R.~S.; McAllester, D.~A.; Singh, S.~P.; and Mansour, Y. 2000.
\newblock Policy gradient methods for reinforcement learning with function
  approximation.
\newblock In \emph{Advances in Neural Information Processing Systems},
  1057--1063.

\bibitem[{Tang(2018)}]{tang2018actor}
Tang, L. 2018.
\newblock An actor-critic-based portfolio investment method inspired by
  benefit-risk optimization.
\newblock \emph{Journal of Algorithms \& Computational Technology} 12(4):
  351--360.

\bibitem[{Tavakoli, Pardo, and Kormushev(2018)}]{tavakoli2018action}
Tavakoli, A.; Pardo, F.; and Kormushev, P. 2018.
\newblock Action branching architectures for deep reinforcement learning.
\newblock In \emph{Thirty-Second AAAI Conference on Artificial Intelligence},
  4131--4138.

\bibitem[{Ye et~al.(2020)Ye, Pei, Wang, Chen, Zhu, Xiao, and
  Li}]{ye2020reinforcement}
Ye, Y.; Pei, H.; Wang, B.; Chen, P.-Y.; Zhu, Y.; Xiao, J.; and Li, B. 2020.
\newblock Reinforcement-learning based portfolio management with augmented
  asset movement prediction states.
\newblock \emph{arXiv preprint arXiv:2002.05780} .

\bibitem[{Yu et~al.(2019)Yu, Lee, Kulyatin, Shi, and Dasgupta}]{yu2019model}
Yu, P.; Lee, J.~S.; Kulyatin, I.; Shi, Z.; and Dasgupta, S. 2019.
\newblock Model-based deep reinforcement learning for dynamic portfolio
  optimization.
\newblock \emph{arXiv preprint arXiv:1901.08740} .

\bibitem[{Zhang and Tao(2020)}]{zhang2020empowering}
Zhang, J.; and Tao, D. 2020.
\newblock Empowering Things with Intelligence: A Survey of the Progress,
  Challenges, and Opportunities in Artificial Intelligence of Things.
\newblock \emph{IEEE Internet of Things Journal}
  \doi{10.1109/JIOT.2020.3039359}.

\end{thebibliography}

\end{document}